# Comparison of Epilepsy Induced by Ischemic Hypoxic Brain Injury and Hypoglycemic Brain Injury using Multilevel Fusion of Data Features

**Sameer Kadem[1], Noor Sami[2], Ahmed Elaraby [3,4], Shahad Alyousif [5,6], Mohammed Jalil[7], M. Altaee[8], Muntather Almusawi[9], Ismaeel, A. Ghany[10], Ali Kamil Kareem[11], Massila Kamalrudin[12], Adnan Allwi ftaiet[13]**

[1] Dijlah University College,Baghdad, Iraq
[2] Department of computer engineering techniques, Mazaya University College, Th Qar, Iraq
[3] Department of Computer science, Faculty of computer science and information, South Valley University,Qena, Egypt
[4] Department Of Cybersecurity, College of Engineering and Information Technology, Buraydah Private Colleges, Buraydah, Kingdom of Saudi Arabia
[5] Department of Electrical and Electronic Engineering, College of Engineering, Gulf University, Sanad 26489, Kingdom of Bahrain.
[6] Research centre, Northampton university, faculty of engineering, department of electrical and electronics engineering, University Drive Northampton NN1 5PH, UK
[7] Department of computer science, Alturath University College, Baghdad, Iraq
[8] Department of medical engineering techniques, Alfarahidi University, Baghdad, Iraq
[9] College of technical engineering, The Islamic University, Najaf, Iraq
[10] Computer Technology Engineering, College of Engineering Technology, Al-Kitab University, Iraq
[11] Department of Biomedical Engineering, Al-Mustaqbal University College, 51001 Hillah, Iraq
[12] Faculty of Information & Communication Technology, Universiti Teknikal Malaysia Melaka, 75450 Durian Tunggal, Melaka
[13] Medical instruments engineering techniques, National University of science and technology, Thi Qar,Iraq

Emails: Sameer.kadhum@duc.edu.iq; noora.sami@yahoo.com; ahmed.elaraby@svu.edu.eg; Dr.shahad.alyousif@gulfuniversity.edu.bh; mohammed.jalil@turath.edu.iq; m.altaa@alfarahidiuc.edu.iq; Muntather.almusawi@iunajaf.edu.iq; ayad.ghany@uoalkitab.edu.iq; ali.kamil.kareem@mustaqbal-college.edu.iq; massila@utem.edu.my; adnan.alameri@nust.edu.iq

## Abstract

The study aims to investigate the similarities and differences in the brain damage caused by Hypoxia-Ischemia (HI), Hypoglycemia, and Epilepsy. Hypoglycemia poses a significant challenge in improving glycemic regulation for insulin-treated patients, while HI brain disease in neonates is associated with low oxygen levels. The study examines the possibility of using a combination of medical data and Electroencephalography (EEG) measurements to predict outcomes over a two-year period. The study employs a multilevel fusion of data features to enhance the accuracy of the predictions. Therefore this paper suggests a hybridized classification model for Hypoxia-Ischemia and Hypoglycemia, Epilepsy brain injury (HCM-BI). A Support Vector Machine is applied with clinical details to define the Hypoxia-Ischemia outcomes of each infant. The newborn babies are assessed every two years again to know the neural development results. A selection of four attributes is derived from the Electroencephalography records, and SVM does not get conclusions regarding the classification of diseases. The final feature extraction of the EEG signal is optimized by the Bayesian Neural Network (BNN) to get the clear health condition of Hypoglycemia and Epilepsy patients. Through monitoring and assessing physical effects resulting from Electroencephalography, The Bayesian Neural Network (BNN) is used to extract the test samples with the most log data and to report hypoglycemia and epilepsy patients non-invasively. The experimental findings demonstrate that the suggested strategy improves accuracy by 95.05% and reduces the error rate to 0.41 when comparing diseases.

## 1. Introduction

Brain hypoxia results from not getting enough oxygen to the brain. Anyone hospitalized for cardiac arrest drowns or suffocates may experience brain hypoxia. [1] Trauma, stroke, and carbon monoxide poisoning are a few potential causes of cerebral hypoxia. For brain cells to function properly, they require a steady flow of oxygen, and the situation is severe. [2] The supply of oxygen to the brain is interrupted by several chronic happenings. Stroke, heart failure, and increased heart rate can prohibit travel to the brain of oxygen and nutrients. [3] The brain requires oxygen for its principal energy supply, glucose. When the amount of oxygen is disrupted, identity is lost around a few seconds, and the brain starts to weaken, lacking ventilation after approximately four minutes. Cerebral anoxia is related to a complete disruption of the oxygen supply to the cortex. [4] Cerebral anoxia is considered brain hypoxia if the oxygen consumption remains partial but insufficient to ensure stable brain activity. The two words are used differently words in education. The brain absorbs a huge amount of energy according to weight and scale. Anoxia is a total lack of amount of oxygen in an organism. Hypoxic describes reduced oxygen supply to a body with constant blood circulation. [5] A huge collapse of oxygen to the brain causes brain injury. Inadequate blood circulation in the brain and reduced blood oxygen (hypoxia) contribute to natural cerebral self-regulation impairment. It treats severe brain damage and hypoxic-ischemic. [6]The bulk of the cases of hypoxic brain injuries is heart failure, neurological injuries, losing consciousness, smoke inhalation, panic, drug treatment poisoning, painkillers, carbon monoxide poisoning, or head trauma. Another severe form of hypoxic brain damage is heart failure [7] .

Compared to brain injury, where physical injuries cause changes in the brain and hypoxia are severe injuries caused by a lack of oxygenation. Hypoxic brain wounds have strokes, but strokes are not the origin of brain wounds of this kind. [8] The oxygen-carrying bloodstream that does not enter the brain can contribute to hypoxic damage known as persistent anoxia, which may cause oxygen starvation induced by strokes. However, certain lung disorders, such as heart arrest or heart function, may occur. [9] Everyone is in danger of brain hypoxia if they undergo an incident where they don't have enough oxygen. In circumstances that take users away from oxygen, one's job or daily activity can increase risk. [10] Cerebral hypoxia is a type of hypoxia that is especially caused by the brain. It is called cerebral anoxia if the brain is completely depleted of oxygen. There are four types of brain hypoxia: diffuse cerebral hypoxia, localized cerebral ischemia, brain infarctions, and worldwide cerebral ischemia, in order of gravity. Long-term hypoxia leads to neuronal damage by apoptosis, which leads to hypoxia. [11] A disease that occurs if the whole brain has no sufficient oxygen supply, but the deficiency is not complete, is hypoxic-ischemic (HI). Although cerebral hypoxia can occur at any age and is a common heart attack condition, hypoxia-ischemia is usually connected with the amount of blood in the neonate resulting from asphyxiation at birthing. The brain requires about 3.3 ml of oxygen per 100 g of brain tissue per minute [12].

Blood diverted to the cortex and through the bloodstream causes the organ to respond immediately to a drop in oxygen saturation. The typical course of blood circulation may increase by almost two times, but not more. But if blood circulation is not enhanced or blood circulation is not increased, cerebral hypoxia diagnoses may occur. [13] Slight effects involve problems with the complicated learning process and short-term memory losses. Cognitive problems and impaired motor function occur if oxygen deficiency persists. [14] Hypoglycemia normally causes a lack of brain oxygen, which leads to functioning brain loss that an increase in the concentration of glucose levels can reverse. [15] Hypoglycemia causes brain fuel starvation that initially activates various defenses, but uncontrolled brain dysfunction outcomes are usually fixed after increased plasma glucose levels. Hypoglycemia causes Hypoglycemia in diabetic people. Hypoglycemia rarely causes intense brain death, and at most, in species extended. Hypoglycemia, particularly in people with diabetes, is a chronic symptom. Hypoglycemia causes common and non-specific neurological symptoms, such as anxiety, hallucinations, focal nerve defectives, haze, and depression. In the case of short-term Hypoglycemia, behavioral defectives are generally completely reversible and semi-life-threatening. Hypoglycemic encephalopathy occurs throughout deep and prolonged Hypoglycemia is a continued brain-dead state. Headaches, as well as other behavioral deficits, could be noted. For individuals who have insulin-treated diabetes, Hypoglycemia is the rate-limiting medication obstacle[26][27][28]. The incidence of serious Hypoglycemia in insulin-treated patients' wide range of networks with greater pressures on insulin sensitivity. On average, once every year, serious, life-threatening Hypoglycemia develops for diabetic chronic treatment. Enhanced distress caused by high blood glucose is induced by the depletion of the brain, which leads to convulsions and harm to the depression and brain. Hypoglycemia-induced brain injury leads to significant cognitive problems with space-related and direction memory shortcomings in clinical and experimental research. Hypoglycemia increases signals that prevent neurons from repolarizing and regulating the flow of calcium. The amount of calcium inflow from either the tubules or cellular membrane space contributes to an important feedback mechanism leading to necrosis of the brain cells. Hypoglycemia typically happens in newborns with descendants of diabetic mothers or with low prenatal birth weight. Many hypoglycemic babies have no signs and symptoms, while some are emblematic and, therefore, at threat of irreversible brain damage. This study employs a hybridized categorization model for brain injury caused by HCM-BI.

Here is the assignment for the remaining work. Part 2 gives background study insights; Part 3 examines the HI results of each infant and applies support vector machines with clinical information. The Bayesian neural network (BNN) is used to non-invasively diagnose hypoglycemia as well as epilepsy patients and extract the test samples with the most log data by observing and evaluating physical impacts brought on by EEG. The results are validated in Part 4. The research is concluded in Part 5.

## 2. Related Works

This section discourses numerous works investigators have done; Johanna C. Harteman et al. [16] examine the association between placental disease and brain injury trends in full-term children with neonatal encephalopathy following a perceived attack. The study consisted of neonatal encephalopathy full-termed babies following suspected hypoxia-ischemia with the placenta to be screened for cerebral MRIs in the first 15 days following delivery. The placenta was examined for both macroscopic and microscopic features. The children were categorized as per the prevalent brain injury results presented in MRI. Clinical factors were reported, both maternal and perinatal.

H. Ma et al. [17] evaluated tests on different depths and hypothermia rates for investigating neuron protective effects were conducted. This review offers an overview of the latest knowledge on the benefits of mechanical ventilation for neonatal hypoxic-ischemic brain damage and brain injuries. It discusses the metabolic mechanisms that may contribute to new treatments. However, this study discusses the physiological efficacy of antibiotic therapy for infant hypoxic encephalopathy and head trauma for adults. Various cellular and molecular problems that cause lead to mental insult are caused by hypoxic-ischemic brain damage and brain injury. Clinical hypothermia activation tends to enhance neuronal death in the cellular situation. Primary injury damage, usually developing oxidative stress, neurons, inflammatory processes, and apoptosis, is attenuated by Hypoglycemia

Dan Shlosberg et al. [18] examined the primary source of humanity in early people: traumatic brain injury (TBI). TBI recovery in the acute phases has vastly enhanced. Though, the long-term effects of treatment and control remain a concern. Blood-brain barrier (BBB) disintegration has frequently been reported in TBI patients, but the significance of dysfunctional vascular disease has been currently examined. The increased proportion of BBB condition has been confirmed by brain imaging evidence in patients with TBI, indicating that biomarkers in the clinic and drug tests may be used as such pathology. In this segment, the neurological implications of TBI with an emphasis on the long-term complications are addressed. The structure facilitates evidence for the existence of BBB in TBI and discusses the primary and secondary processes involved in this condition.

Richard Idro et al. [19] look at potential brain damage processes in cerebral malaria linked to the disease progression and aim to enhance neurobehavioral outcomes. Cerebral malaria is Plasmodium falciparum inflamed appendix's most serious neurological problem. The risk of physical and cognitive disorders, behavioral issues, and seizures that sustain patients are improved. The leading cause of pediatric neuron disability in the area is cerebral malaria. Neurocognitive squeal pathogenesis is misunderstood: coma evolves across different mechanisms, and many methods of brain damage can occur. The cause of such brain damage is unknown as an intravascular infection. To develop effective neuron protective therapies, recognizing these processes is critical.

Hesham Montassir et al. [20] reported that patients with systemic occipital lobe due to prenatal Hypoglycemia commented on their long-term clinical path. Six neonatal and occipital lobe epilepsy patients were evaluated at our thorough hospital examination of the health records. The age of average epilepsy was two years, eight months, 12 years, and four months for median follow-up. The original seizure types were four patients with generalized seizures. The key seizures were similar to occipital lobe seizures, such as divergence of the limb, limb swelling, vomiting, and visual hallucination. During the early stage of epilepsy, both patients had epileptic symptoms. For both cases, EEGs revealed parietal-occipital peaks. In 4 patients, RIM reported parietal-occipital cortical atrophy and T2 extension in one hippocampal atrophy and rare in one patient.

Hongmei Song et al. [21] deal with the middle brain artery in adult black (C57). The hippocampal and neocortical areas by mouse and electroencephalographic (EEG) were tracked. Electrographically separations were found within 90 min of the center brain artery deformation in the lack of explosive motor behaviors. Hippocampal releases are much more intense than related cortical flushes in seizure cases studied. Hippocampal flushes were found in certain seizure cases alone or with limited cortical intervention. As established by corresponding cytological exams, hippocampal damages were correlated with seizure growth. Hypoxia-hypoglycemia events were implemented in the mouse

Using neuronal Fyn overexpressing (OE) mice, Renatta Knox et al. [22] examines the impact of neuronal Fyn in neonatal brain HI. HI was detected using the Vannucci method on a postnatal day 7 in wild-type (WT) and Fyn OE mice. Five days later, cresyl violet and iron stains were evaluated to determine the extent of the damage. Western blotting of synaptic membrane proteins and co-immune supporting proteins with postsynaptic density (PSD) was carried out at various time periods following HI to determine the phosphorylation and activity of NMDA receptors. Comparing Fyn OE mice to their WT littermates, we can see that they had significantly higher mortality and brain damage.

Dipti Kapoor et al. [23] documented the electro-clinical range of children with Neonatal hypoglycemic brain injury (NHBI) secondary epilepsy is documented. It was a secondary analysis of school children in the Clinic for Epilepsy from January 2009 to August 2019. The samples were recorded of children who experienced epilepsy following reported neonatal simply symptom hypoglycemia. Description of the clinical history, type of epilepsy, co-morbidity of neural development, neurology, and seizure results is recorded. The research focuses on a diverse electric and radioactive materials spectrum and the harmful effects of NHBI on epilepsy and neurobehavioral.

Jun Su et al. [24] discuss the physical change of pathology and pathogenicity of the NHBI. With infant brain hypoglycemic damage and analyses the relevant features. The "asymptomatic" or "symptomatic" Hypoglycemia procedure are problem relating to the threshold values of NHBI. Continuous or chronic Hypoglycemia can cause permanent neonatal brain trauma and lead to cognitive decline, vision problems, epilepsy of the occipital lobe, cerebral paralysis, and other complications. Many physicians are yet to recognize NHBI. Due to the absence of clinical conditions, there is no specific purpose for guiding NHBI, but the study of brain scans has now become an essential descriptive and prognostic method.

Hannah C. Glass et al. [25] examine the prevalence of epilepsy and health risks in those diagnosed with neonatal long-term acute neurology in primary jurisdiction seizures. Epilepsy risk was smaller and older when it occurred than in earlier studies; several causes may be affected, including neuro crimes, hypothermia management of HIE, high levels of neonatal transfer to nursing care, and the absence of neonatal-onset epilepsy. The risk of epilepsy was not decreased by more anti-seizure treatment during childhood. Multicenter investigations are necessary over a lengthy period to determine if neonatal seizure can change the epilepsy threat.

Based on the statistical survey, HCM-BI is used to find the HI outcomes of each infant; the support vector machine is applied with clinical details. Through monitoring and assessing physical effects resulting from EEG, the Bayesian neural network (BNN) is used to diagnose Hypoglycemia in a non-invasive way as well as epilepsy patients and extract the test samples with the maximum log data.

## 3. Hybridized classification model for Hypoxia-Ischemia and Hypoglycemia (CMHI-H) brain injury

These brain injury diseases are based on neuropathology findings of the hippocampus and cerebral line and the assumptions of specific neuronal necrosis due to energy loss HI, Hypoglycemia, and increased energy demand due to epilepsy.

### 3.1. Support Vector Machine to identify the HI outcomes of newborn babies

The primary cause of neurobehavioral illness in children appears to be the hypoxic and ischaemic damage that occurs in neonatal periods. About 3% of every thousand live births are assumed to affect this condition, which is significantly important because of the advancement of adequate preventive treatments. Doctors focus exclusively on actions to assess fetal growth throughout delivery and predict prognosis. Cardiotocography (CTG) is an essential method for fetus tracking. Many babies with HI are found to be functioning with regular Cardiotocography. Even during the study, encouraging behaviors of the fetus's heartbeat are missed, and abnormal or diseased trends for Cardiotocography are created. Appearance, Pulse, Grimace, Activity, and Respiration values are uniformly given to newbies at birth. This measuring instrument attempts to know the state of the infant immediately after birth. If electroencephalography is usual in the first eight hours, it is said that it is an outstanding prediction for a positive result. The electroencephalography operation remains dormant or extremely irregular, and also bad results of electroencephalography signals. An advanced diagnosis is extremely beneficial for accurately modifying clinical procedures in impaired newborns and informing families and clinicians. The multisensory evidence to analyze the potential neurobehavioral result of the infant. Since the Cardiotocography information was not accessible, the postnatal heartbeat data was collected at 12 hours. This information is transferred to a support vector machine to learn learning patterns for a good or bad performance.

#### 3.1.1. Data collection

Babies who met two or all of the following circumstances are included in this study Appearance, Pulse, Grimace, Activity, Respiration values or irregular neurological, medical epilepsy, and babies' Electroencephalography data. Children that fulfilled the basic requirements are tested using a traditional pediatric neural evaluation process. The infants with the replacement of 10-20 electrodes are used, and EEG data are recorded. The recordings are obtained from a particular hospital. Electroencephalography data are recorded every 12 hours.

### 3.1.2. Retrieval of certain features

Some important features are considered in the extraction stage (a) Operator of non-linear systems (ONLS), and it is defined in equation (1)

$$ONLS[j] = u^2(j) - u(j-1)u(j+1) \quad (1)$$

Here $u$ is the Electroencephalography data, and it is more reliable to changes in intensity as well as magnitude, and it is shown in figure1. (b) The average root squared value of Electroencephalography data, (c) the total Electroencephalography spectrum energy in the group, (d) Power separated by actual strength in the delta frequency range.

3.1.3. Differentiation of the samples

In this case, take into consideration a training dataset marked as $(a_j, b_j), j = 1, \ldots . m, b_j \in [-2,2], a_j \in S_b$.

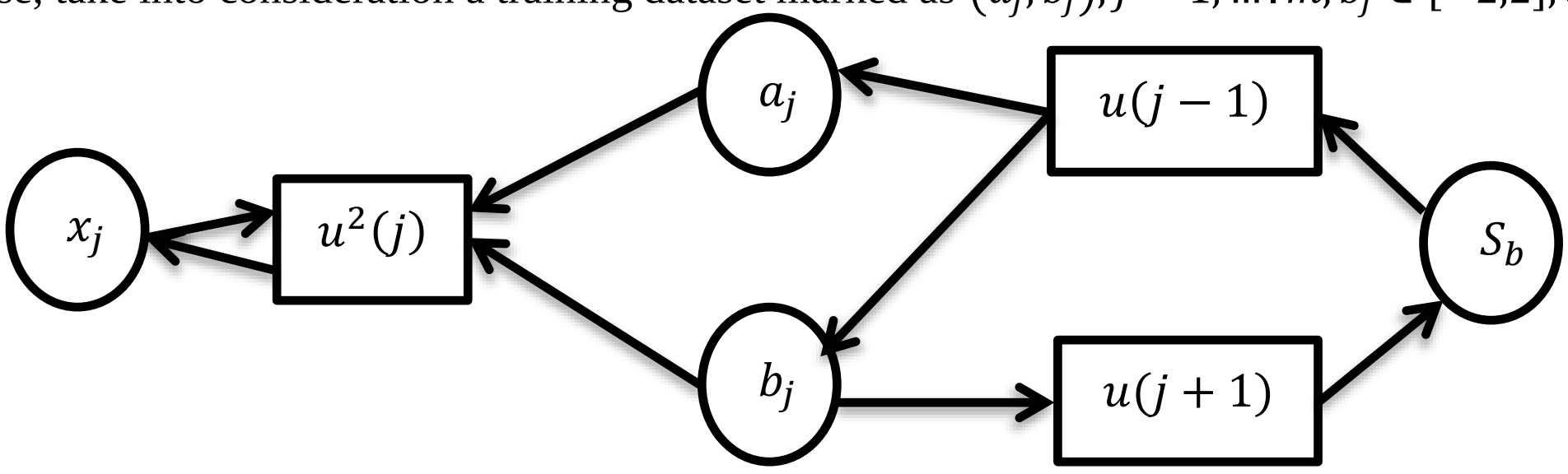

Figure1: Non-linear system and its delay

The Support Vector Machine approach defines the higher dimensional space, which holds both theoretical loss low and optimizes the distance between the planes and the nearest cases. It is done numerically, and it is shown in equations (2) and (3)

$$red(\frac{1}{2}\|y\|^2 + D\sum_{j=1}^{m}\epsilon_j) \quad (2)$$

$$x_j(\delta.\varphi(y_j) + C) \geq 1 - \epsilon_j \quad (3)$$

$$\epsilon_j \geq 0\forall_j$$

Here $\delta$ is usual to a vector space, the variables to warm the judgment limits $\emptyset$, and the non-linear activation method of the input room to the function room to are weak parameters. The $d$ factor is a positive number of normalization that governs the distinction between margin maximization and error minimization. For example $a$, the category forecast is given by the equation (4)

$$g(a) = \sum_{i=1}^{N}\beta_i b_i\langle\emptyset(b_i).\emptyset(a)\rangle + d \quad (4)$$

$N$ is supported by the total amount of matrices and $\beta_i$ constants are measured by Lagrangian maximum, and it is shown in equations (5) and (6)

$$\sum_{j=1}^{m}\beta_j - \frac{1}{2}\sum_{j=1}^{m}\sum_{n=1}^{m}\beta_j\beta_n b_j b_n\emptyset(a_j).\emptyset(a_n) \quad (5)$$

$$\sum_{j=1}^{m}\beta_j b_j = 0 and\ 0 \leq \beta_j \leq D, \forall_j \quad (6)$$

The positions $\beta_j > 0$ are known as the training examples and are the nearest to the feature space. A kernel can allow the non-linear transformation of a, and it is shown in equation (7)

$$H(a_j, a_n) = \langle\emptyset(a_j).\emptyset(a_n)\rangle \quad (7)$$

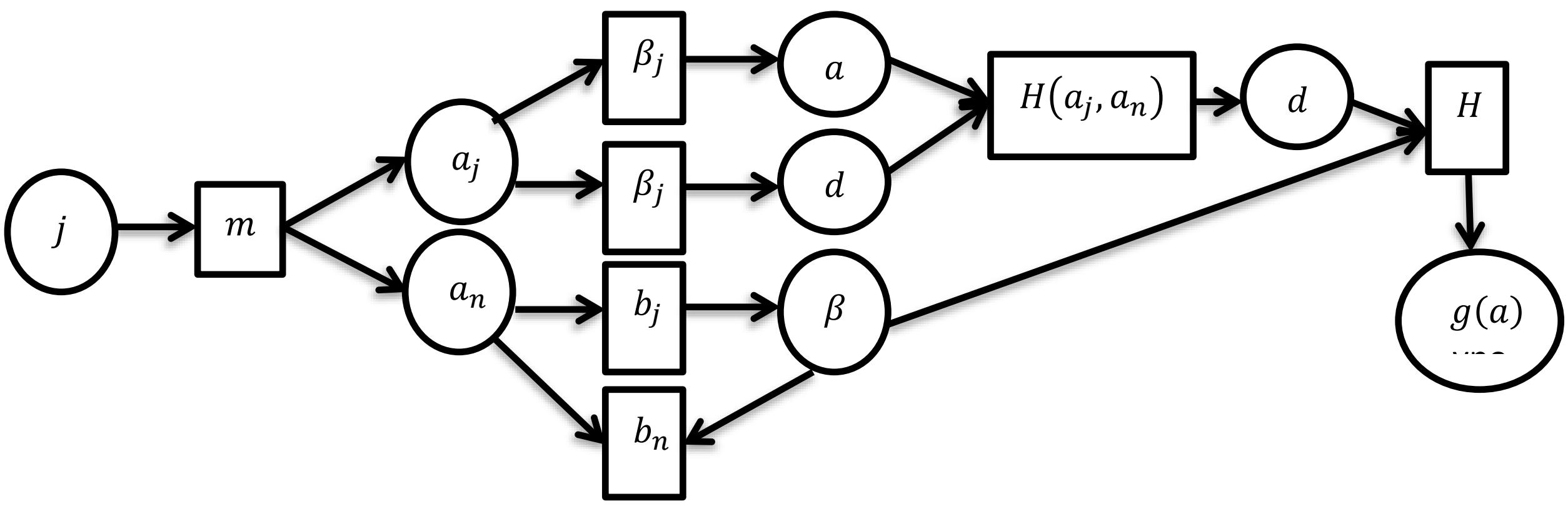

Figure2: SVM implementations are the non-linear activation function

Here $H$ is the kernel of the internal product. One of its most widely utilized classification algorithms in SVM implementations is the non-linear activation function, and it is shown in equation (8) and illustrated in figure2

$$g(a) = \sum_{n=1}^{m}\beta_n b_n H(a_j, a_n) + d \quad (8)$$

### 3.1.4. Support vector machine execution

Both periods during the practice are numbered 2 and -2, respectively, for children with usual and unusual results. Each function is removed, and the Classification method is fed to the period. The training samples for the Support Vector Machine classification are first cylindrically normalized by eliminating the medium and separating the different characteristics by a normal distribution. Apply this standardizing framework to the test results. Cross-validation (CV) is used in these projects to conduct an objective performance evaluation to guide information across all but a single person while checking for the remainder of the physician. It is replicated before each patient has been checked, which helps to reduce the error rate. Nestled CV is implemented in addition to the learning data to try for appropriate classification parameters. After selecting the right variables, the end system is built overall dataset. Support Vector Machine results are transformed by the sigmoid method to deterministic steps as shown in equations (9) and (10)

$$Q\left(b = \frac{1}{h}\right) = \frac{1}{1+\exp\,(xh+y)} \tag{9}$$

$$Q\,(x|y) \propto Q(x)\sum_v Q(y|v)\sum_d Q(d)Q(v|d,x) \tag{10}$$

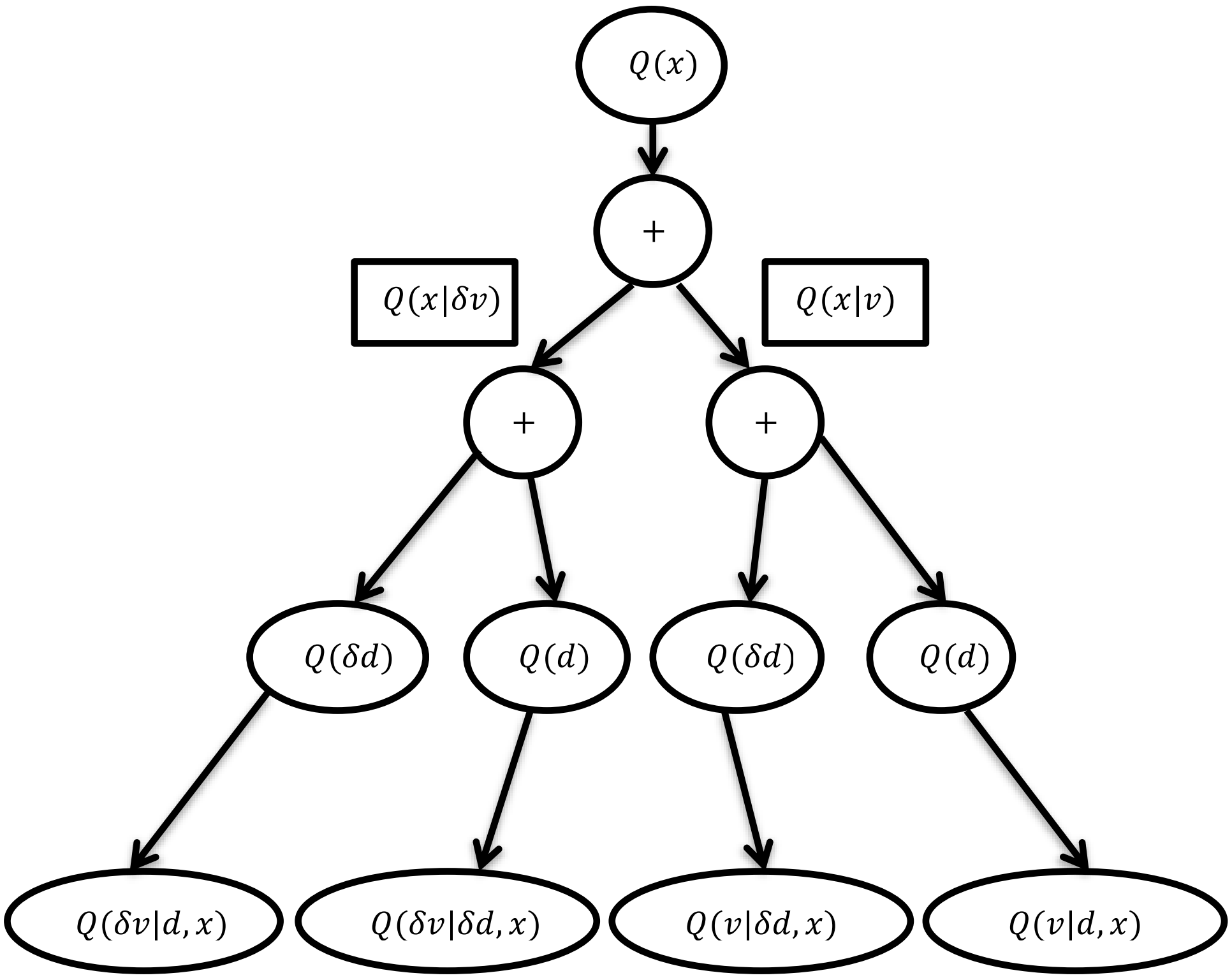

Figure3: Support Vector Machine predicted results

Here h is a Support Vector Machine performance, $x$ and $y$ are the matrix multiplication-function variables, calculated by the process reported on the testing set. Support Vector Machine production by multiplying, whereby b is the Support Vector Machine production, and the lowest value corresponding is said to increase the Support Vector Machine output substantially about the projection of predicted results is shown in figure 3.

3.1.5. Integration of criterion and judgment

A level for the stochastic indicators measured is implemented to allocate a category at each iteration. This level is determined based on the dataset per person. The final Support Vector Machine is translated into stochastic steps and contrasted with the basic facts. The limit for a terminal differed from 0 to 1 in stages for calculating the area of the receiver operating function. The maximum performance level for the CV is chosen and added directly to the study analysis to determine the reliability of the prediction per person. Each person made several judgments in the Support Vector Machine by one result. The maximum voting system incorporated such a process for patient-based evaluation of outcomes.

### 3.2. Bayesian neural network (BNN) to diagnose Hypoglycemia

BNN is used to diagnose Hypoglycemia in a non-invasive way. Few currently offered regular blood glucose factors are impacting, with different working, price, performance, and overloading disadvantages. The production of Hypoglycemia uses a variety of concepts from the identification of mood changes from sweating to insulin sensory data of blood glucose in surrounding tissues. None of this has, however, been accurate or discreet enough. Glucose monitoring devices now provide real-time blood glucose estimates; they cannot always be used as warnings, particularly in the hypoglycemic context. With most patients with type 1 diabetes, Hypoglycemia is a normal part of life. Hypoglycemia events are extreme symptoms common to the person, frequently causing an attack or a collapse. The related incidence, function loss, and disability risk worsen Hypoglycemia for patients and healthcare professionals. A healthy supply of sugar is essential to the brain and is susceptible to a lack of insulin. The intellect is the first body to which blood sugar levels and cannot formulate or retain these energy sources. Hypoglycemia is produced when neuronal insulin release by the circulatory system decreases glucose intake. The integration of several protective factors usually corrects Hypoglycemia. A range of options can arise from moderate and severe seasons of Hypoglycemia.

Mild perspiration, uneasiness, heart pumping, confusion, anxiety, etc., rarely causes perspiration. It can be determined by consuming or consuming nutrition rich in sugar. Hypoglycemia could become serious and contribute to seizures, coma, and even failure if left unchecked. Hypoglycemia lowers victims and their standard of living by triggering persistent anxieties regarding the possible prospects of Hypoglycemia. Hypoglycemia unconsciousness is one of the harmful effects of Hypoglycemia. It is because Hypoglycemia often happens, which can affect patients' reactions. In circumstances of knowledge, the organs of clients do not emit the substance insulin that triggers clinicians with early symptoms such as sweat, starvation, and agony. Any alert usually prevents patients from understanding Hypoglycemia until everything is extreme and can lead to serious injury. Originally, due to insulin resistance, decreased blood sugar levels are decreasing. With a continuous decrease in sugar levels, unnecessary balance-regulatory factors at certain levels are sequentially accessed to guarantee adequate glucose consumption in the brain. In Type 1 diabetes ( T1D), collapsing blood glucose does not generate insulin secretion reactions at standard glycemic control limits that enable the blood glucose to drop into harmful levels of Type 1 insulin treatment. The glucagon anti-regulatory reaction to Hypoglycemia would be ineffective from the first couple of years of a person with diabetes.

Moreover, warning signs are no longer present in some cases, and the condition can lead to severe problems such as coma and seizures. This effect is called Hypoglycemia. Throughout Hypoglycemia, stimulating the central nervous system is the most serious biological adjustment. Sweating and enhanced cardiovascular activity are some of the worst reactions. Body sweating is induced by protective neuronal bacteria, whereas the changes in heart rate and blood pressure are due to increased heart rate.

#### 3.2.1. BNN for Hypoglycemia

The Bayesian network and Gaussian weight derivatives achieve neural networks with the multi-layers, which can offer the most significant widespread effect. The weight training w of network X, in particular, is modified following the training data D to the most likely values. The loads $\omega$ of system Y, in general, are set with their most likely values in light of training sample B. In particular, the longitudinal mass distributions can be determined according to Bayes' law and is shown in equation (11)

$$q(\omega|B,Y) = \frac{q(B|\omega,Y)q(w|Y)}{q(B|Y)} \quad (11)$$

Here $q(\omega|B,Y)$ is the probability function, containing analysis weight information, and the $q(w|Y)$ normal probability includes context-awareness weight information. The $q(B|Y)$ is the denominator, called system Y proof.

Regulations could be used to discourage unnecessary loads that can result in bad widespread use. A load decay condition that is associated related to the number of ranges of the masses and partitions of the H group neural network category is based on the data error signal $S_E$ to provide the objective functions for a multi-layer network is shown in equation (12)

$$T = S_E + \sum_{h=1}^{H} \varepsilon_h S_{L_h},\ S_{L_h} = \frac{1}{2}\|L_h\|^2\ (h = 1, \dots..H) \quad (12)$$

If the expense feature is called T, the value of a division between weights and biases in the function $L_h$and $\varepsilon_h$ is known as a maximum variable, and the weight and bias vectors for the function are known as $h$. The number of layers used and the weight of the vectors can be represented as $[P_1, \dots.P_6]$ and the transition stage of each vector is stated as $[T_1, \dots.T_4]$ and it is shown in figure4

.

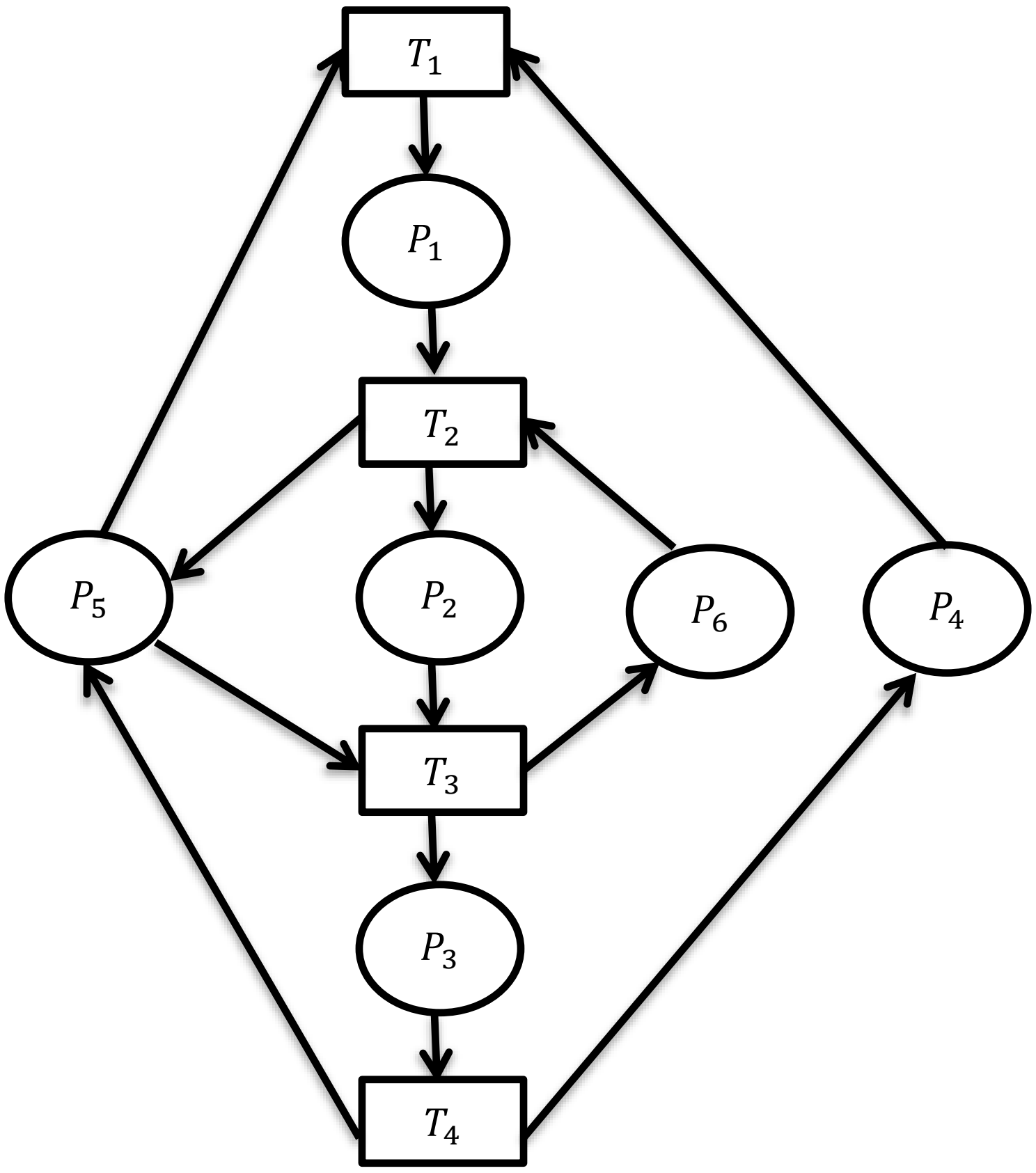

Figure 4: The transition stage of a vector

The input variables must be multiply recalculated until another cost function respect considerable adjustments in consecutive years. After completion of the network preparation, the $\beta_h$ and $\varepsilon_h$ input parameters are used as follows to evaluate the log evidence proof of a $Y_j$ a system with N hidden nodes, and it is shown in equation (13)

$$logS_u(Y_j) = -T + \sum_{h=1}^{H} \frac{X_h}{2} log\varepsilon_h - \frac{1}{2}|A| + logN + N\ log2 + \sum_{h=1}^{H} \frac{1}{2}\left(\frac{4\pi}{\beta_h}\right) - H \log\left(log\Omega\right) \quad (13)$$

Here $X_h$ is the category h, and $\Omega$ is fixed to $10^3$The number of connection weights is represented as h. The maximum log proof is used to identify the best network

3.2.2. Optimization by BNN and feature extraction

Based on the duration and frequency domain, differential equations, equilibrium, and other approaches, the key characteristic removal chooses the optimization for the retrieved characteristics. An optimization algorithm is used, and the extracted features are given into BNN to get the perfect classification results for Hypoglycemia and Epilepsy patients. Epilepsy is a chronic infection with short-term brain activity due to unexpected dissolution in an area of the brain. Due to the continuous extreme muscular deformation causing muscle rigidity, the person's condition becomes stiff. It cannot regulate the free width of the muscle, which shows regular exercise and continuous convocations at attack time. The seizure can cause partial brain damage. It can be unexpectedly unable to regulate its actions in all circumstances and locations, causing scalding, flooding, high-site dropping crash, traffic accident, and other dangerous conditions. The disease wounds to spiritual levels, but not severe. In short, it greatly affects people's good life and kills them seriously. Epilepsy is, therefore, the respected neutral system disease in several regions. Even though it has no strong recommendation for severe illnesses, the physician, in a short time, the rapid events by shock have an enormous effect on the body, brain, and understanding of the patient. Actual identification and forecasts of epileptic seizures are highly significant; cases can be properly treated and avoided.

EEG is a complicated sign that provides the individual with unpredictability, non-linear effects, and other characteristics, while work on epilepsy is an essential task in neutral immune to diseases. Accordingly, the proposed method uses various feature extraction methods, allows selection optimization with an optimization algorithm for optimization, and trains the epilepsy dataset with a BNN. The findings confirm that the proposed framework achieves the maximum classification impact. Over which the techniques for extraction primarily used in this paper for feature extraction dependent on the time domain include: the root mean square value (RMS), absolute value integrity (IAV), absolute mean value slope (MAVS), and zero-cross (ZC). BNN consists of a neutral input layer, two hidden layers, and an output layer. In BNN, each layer might be the cartographic feature of the cached surface array, with a core function in progressive similarity and amplification patterns from both ends. The training dataset is denoted as $y_m = [y_{m1}, y_{m2}, \ldots .. y_{yN}]$ and the related tag is denoted as [m=1,2,…,M] and the output layer is denoted as $Z_m = [m = 1,2, \ldots .. M]$. $\varphi(Y, S_j)$ is the output of the jth secret layer $S_j = [S_{j1}, \ldots S_{jN}]$ , j=[1,…J]. The four attributes or features are shown in figure 5 $g_1, g_2, g_3, g_4$

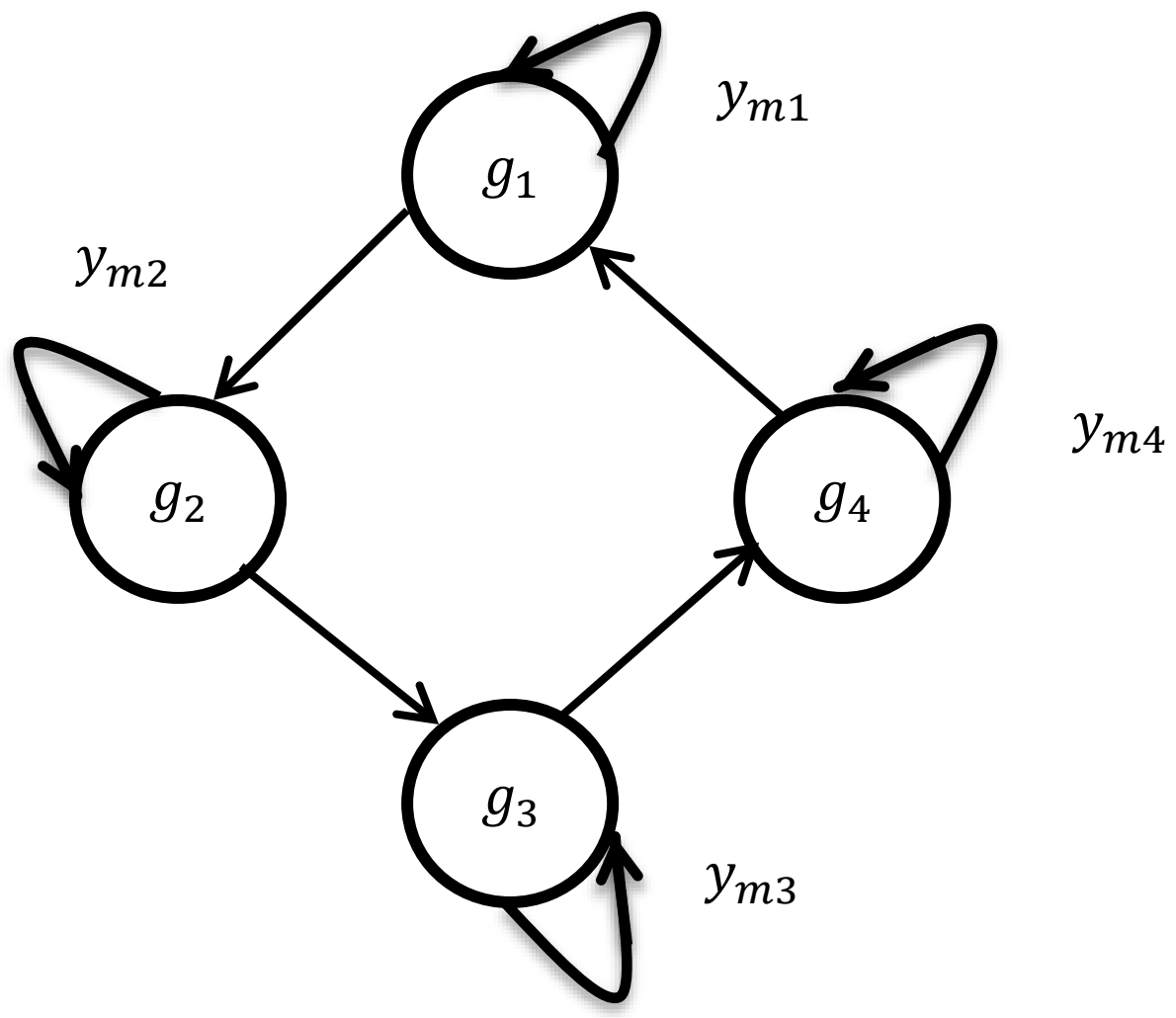

Figure 5: The four attributes with the label data

The main aim would be to find a set of extracted features for each sample. four typical attributes are involved in this stage that can be denoted as $[g_1, g_2, g_3, g_4]$ helps to reduce the accuracy. The first attribute is denoted as $g_1$ root mean square value, and it is given in equation (14)

$$g_1 = \sqrt{\frac{1}{m} \sum_{s=1}^{m} y_s^2} \quad (14)$$

Here s represents the number of secret layers, and y denotes the number of training samples. The second attribute is absolute value integrity, and it is defined in equation (15)

$$g_2 = \sum_{s=1}^{m} | y_s | \quad (15)$$

The third attribute MAVS is shown in equation (16), which refers to the variation between a mean absolute value in the measurements of each wave.

$$g_3 = \frac{1}{m} \sum_{s=1}^{m} (|y_{s+1}| - | y_s | \quad (16)$$

The last attribute is ZC, shown in equation (17), which is used to find the lowest frequency range in EEG samples.

$$g_4 = \sum_{s=1}^{m} SG(-y_{s+1} . y_{s+1}) \quad (17)$$

In this $SG(y) = \begin{cases} 1 \; y \geq 0 \\ 0 y \leq 0 \end{cases}$. The proposed method involves the above four attributes in the feature extraction process.

3.2.3. Classification by BNN

The EEG samples investigation first extracts the epileptic signal feature, selects the optimization by thinking creatively, and makes models by choosing the classifier.

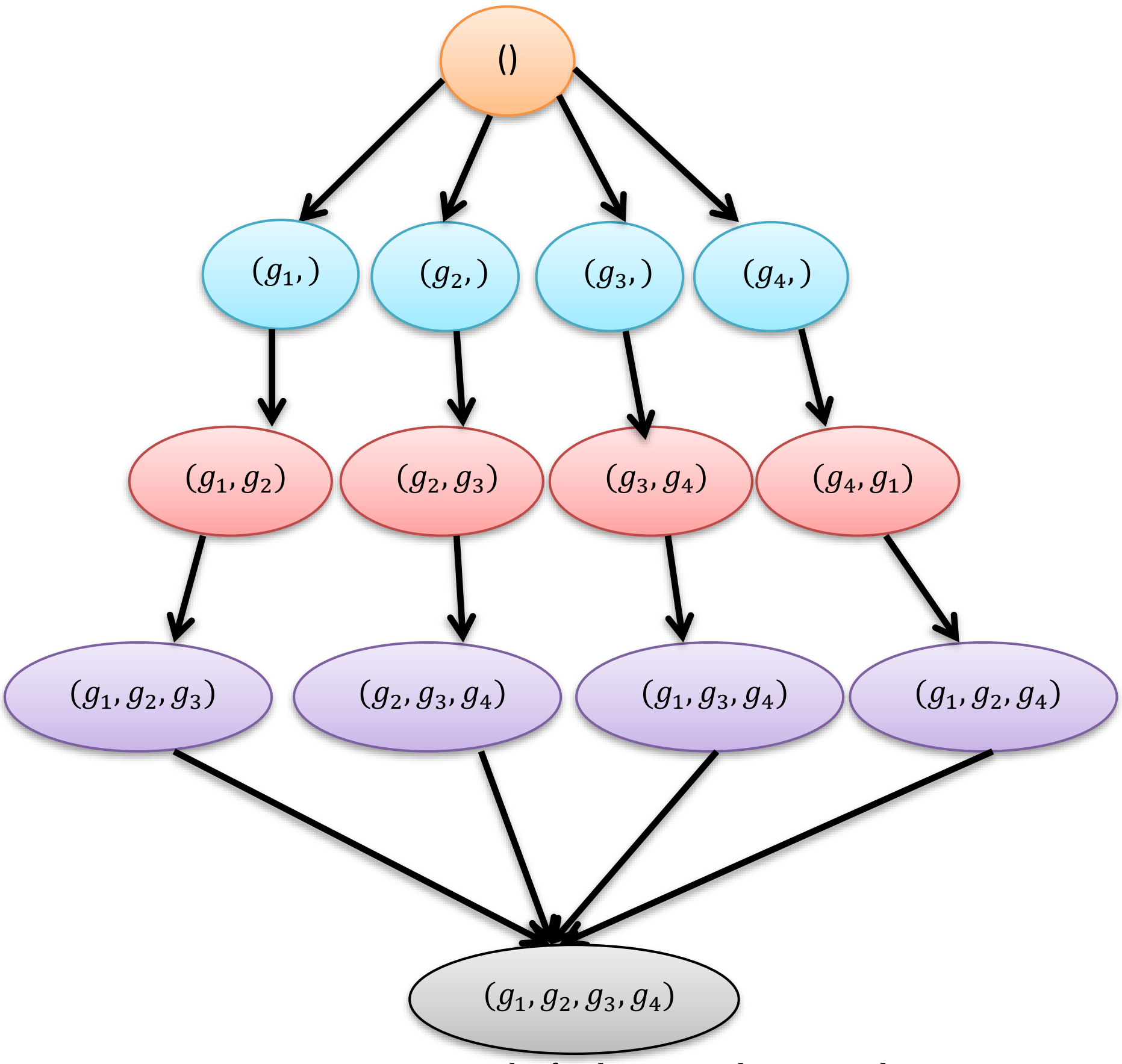

Figure 6: The final stage prediction result

Enter the actual data, select the function, and add a tag to the data set. Insert the derived function variation, along with the unique location and speed, and enter the initial optimization algorithm for health measurement. The optimization algorithm measures the optimal value for each sample and evaluates the best samples accordingly. If good ones exist, substitute them. The acceleration and location of samples are modified and balanced. Start making classification models in training with BNN for the optimum level function sub-set earned from optimization and outcome of the particular model. Enter the data to determine the basis of the BNN classification. The final stage is the prediction of the output result, shown in figure6.

Based on the mathematical equation, the selection of four attributes is derived from the Electroencephalography records, and SVM does not get clear results regarding the classification of diseases. The final feature extraction of the EEG signal is optimized by the Bayesian neural network (BNN) to get the health condition of Hypoglycemia and Epilepsy patients.

## 4. Results and discussions

In this section, HCM-BI has been validated by the database of the Hypoxia-Ischemia (Figure.7(A)) and Hypoglycemia (Figure.7(B)). Epilepsy brain injury affected patients from the hospital as shown in Figure.7(A) and (B)(Case courtesy of Dr Paresh K[redacted] Desai, Assoc Prof Frank Gaillard, Dr Mohammad A. ElBeialy, Dr Nikos Karapasias, Dr Anar Kazimov, Radiopaedia.org,). A major obstacle to improved glycemic regulation in insulin-treated patients is Hypoglycemia, and low oxygen level leads to the results in neonates with HI brain disease. The newborn babies are assessed every two years again to know the neural development results. A selection of four attributes is derived from the Electroencephalography records, and SVM does not get clear results regarding the classification of diseases. The final feature extraction of the EEG signal is

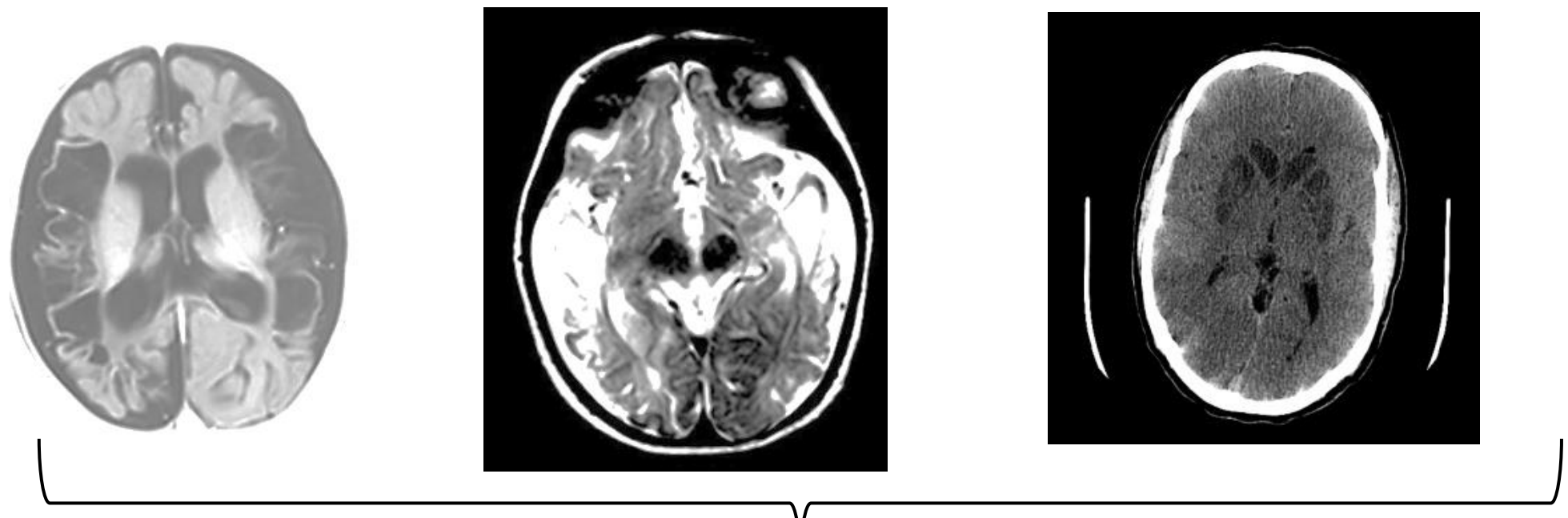

Figure7: (A) (A) EEG pattern, (B) Hypoxia-Ischemia,(C) Changes in Amplitude( Slowing and Increasing)

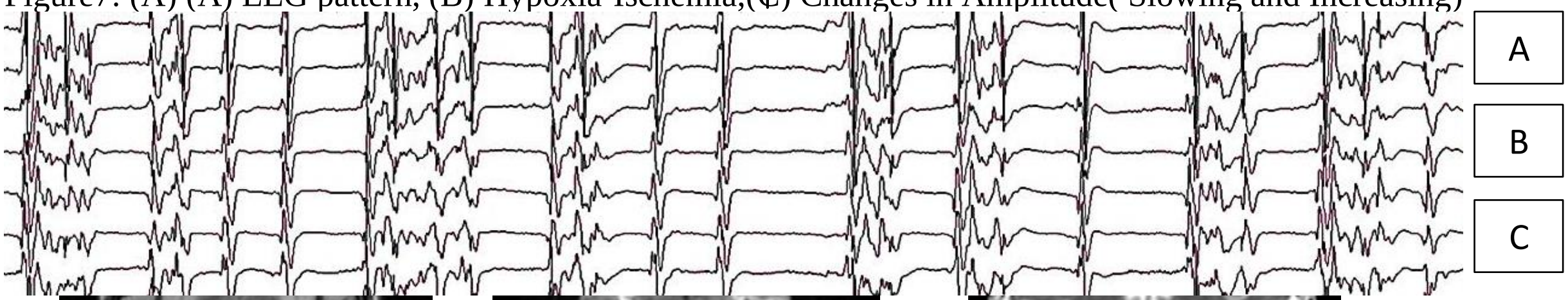

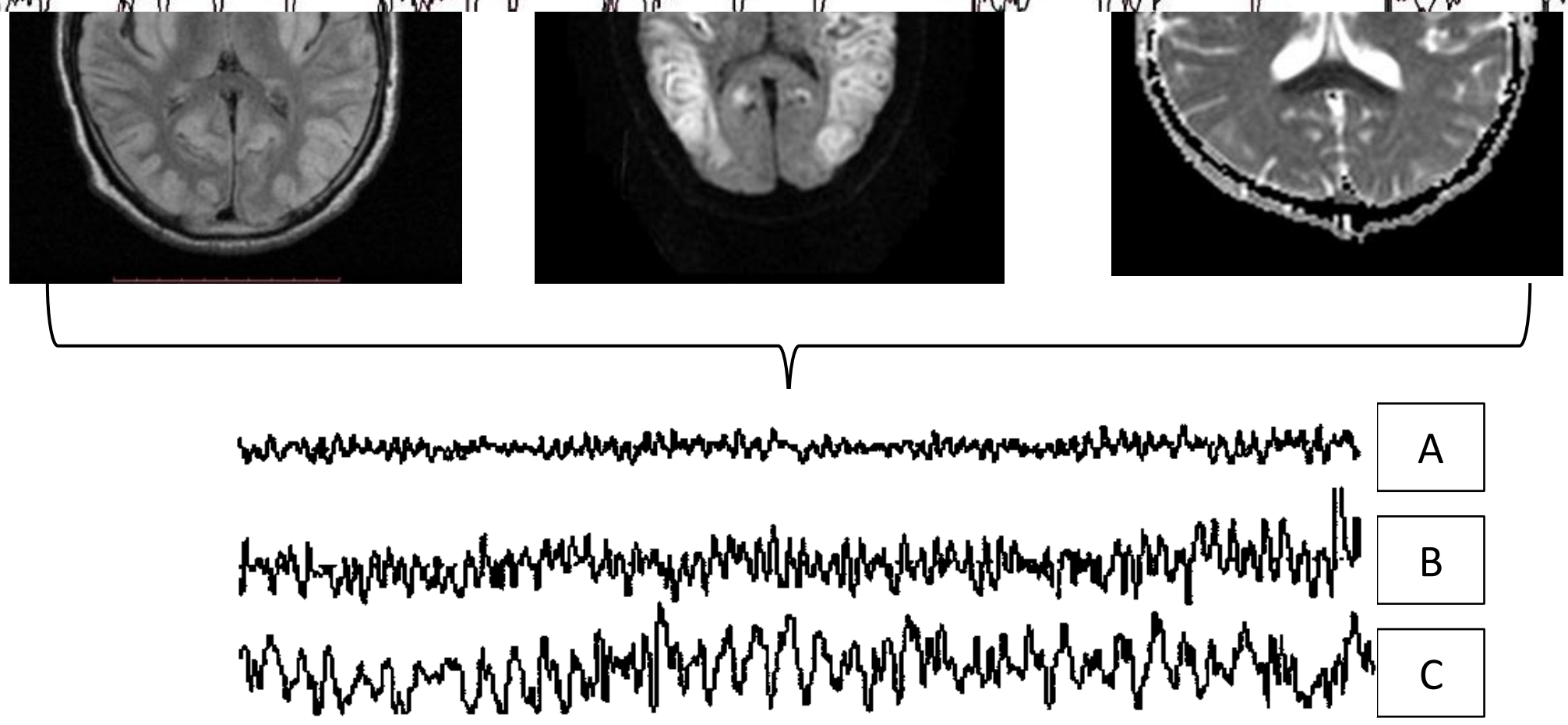

Figure 7(B): (A) EEG pattern, (B) Hypoglycemia,(C) Changes in Amplitude( Slowing and Increasing)

Figure.7. Case study images of Hypoxia-Ischemia (Figure.7(A)) and Hypoglycemia with EEG signal (Figure.7(B)),

They are optimized by the Bayesian neural network (BNN) to get the clear health condition of Hypoglycemia and Epilepsy patients. The Bayesian neural network (BNN) is used to report Hypoglycemia in a non-invasive way as well as Epilepsy patients and extract the test samples with the maximum log data. The classification accuracy for the proposed method is shown in figure8.

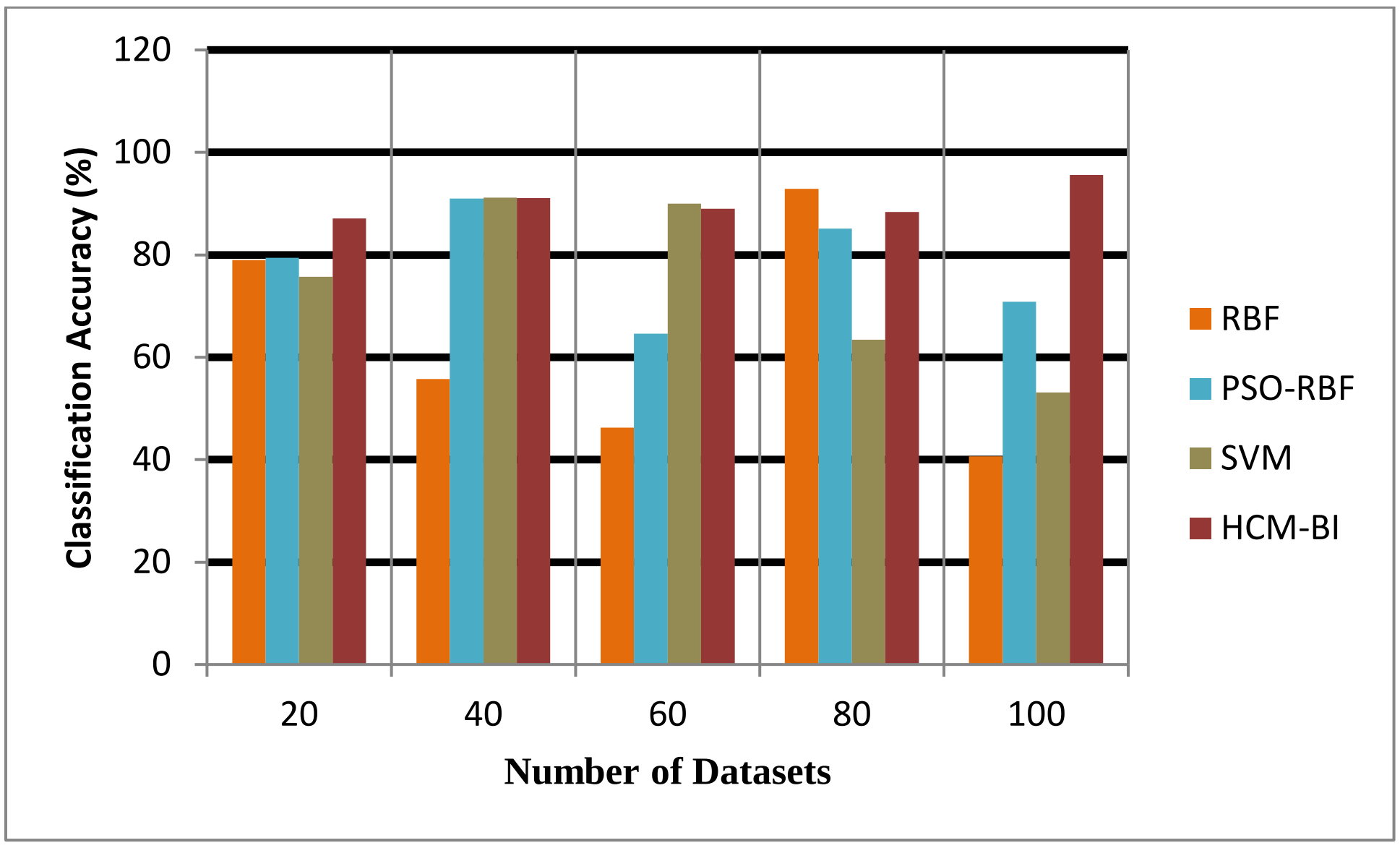

Figure 8: Classification Accuracy of HCM-BI

Hybridized classification model for Hypoxia-Ischemia and Hypoglycemia, Epilepsy brain injury (HCM-BI) could be used to discourage unnecessary weight that can result in bad widespread use. A weight decay condition that is associated related to the number of ranges of the masses and partitions of the neural network category is based on the data error signal $S_E$ to provide the objective functions for a BNN. The error rate of HCM-BI is shown in figure9. Cross-validation (CV) is used in these projects to conduct an objective performance evaluation to guide information across all but a single person while checking for the remainder of the physician. It is replicated before each patient has been checked, which helps to reduce the error rate. Nestled CV is implemented in addition to the learning data to try for appropriate classification parameters.

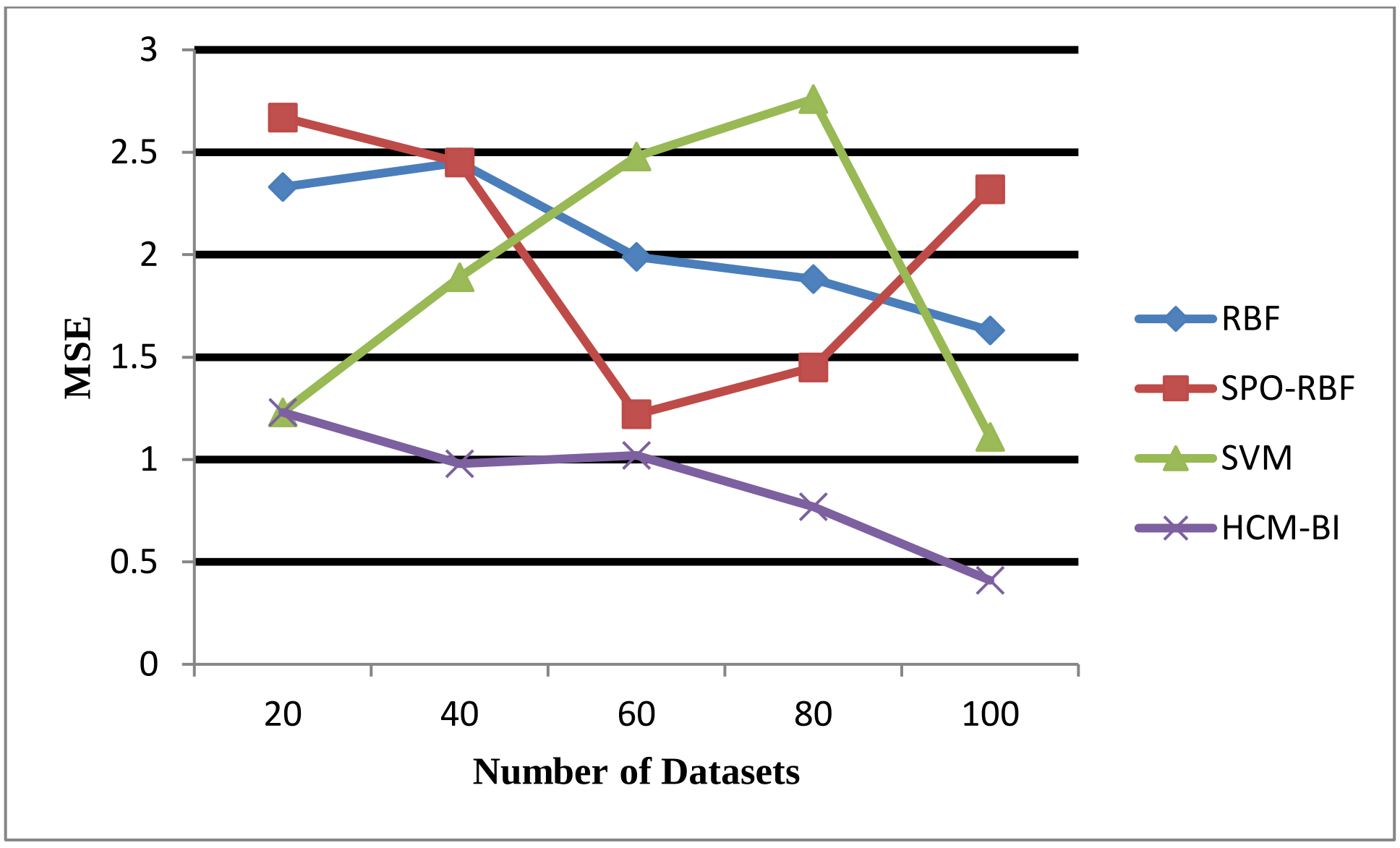

Figure 9: The error rate of HCM-BI

The sensitivity rate of the proposed method is shown in figure10. SVM is used to identify the HI type of diseases in infants, and the number of layers used in processing depends on the sensitivity rate. The sensitivity rate for each type of iteration varies based on the four attributes derived from Electroencephalography records.

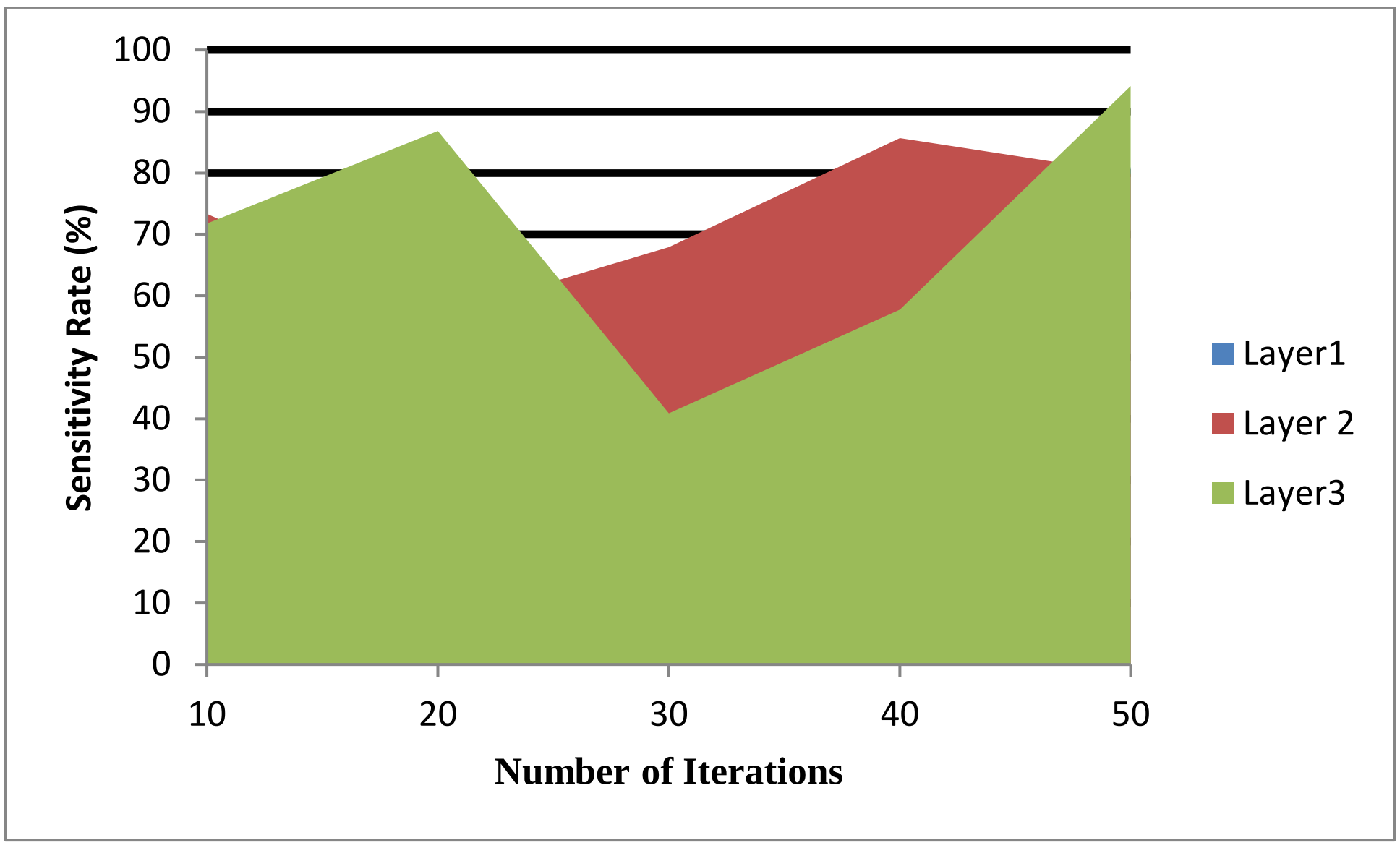

Figure 10: Sensitivity rate of HCM-BI

Integral Absolute Value is one of the attributes that BNN extracts. The second attribute is the Integral Absolute Value, defined in equation (15). The Integral Absolute Value rate achieved by HCM-BI is shown in figure11. Over which the techniques for extraction primarily used in this paper for feature extraction dependent on the time domain include: the root mean square value (RMS), absolute value integrity (IAV), absolute mean value slope (MAVS), and zero-cross (ZC). BNN consists of a neutral input layer, two hidden layers, and an output layer. In BNN, each layer might be the cartographic feature of the cached surface array, with a core function in progressive similarity and amplification patterns from both ends.

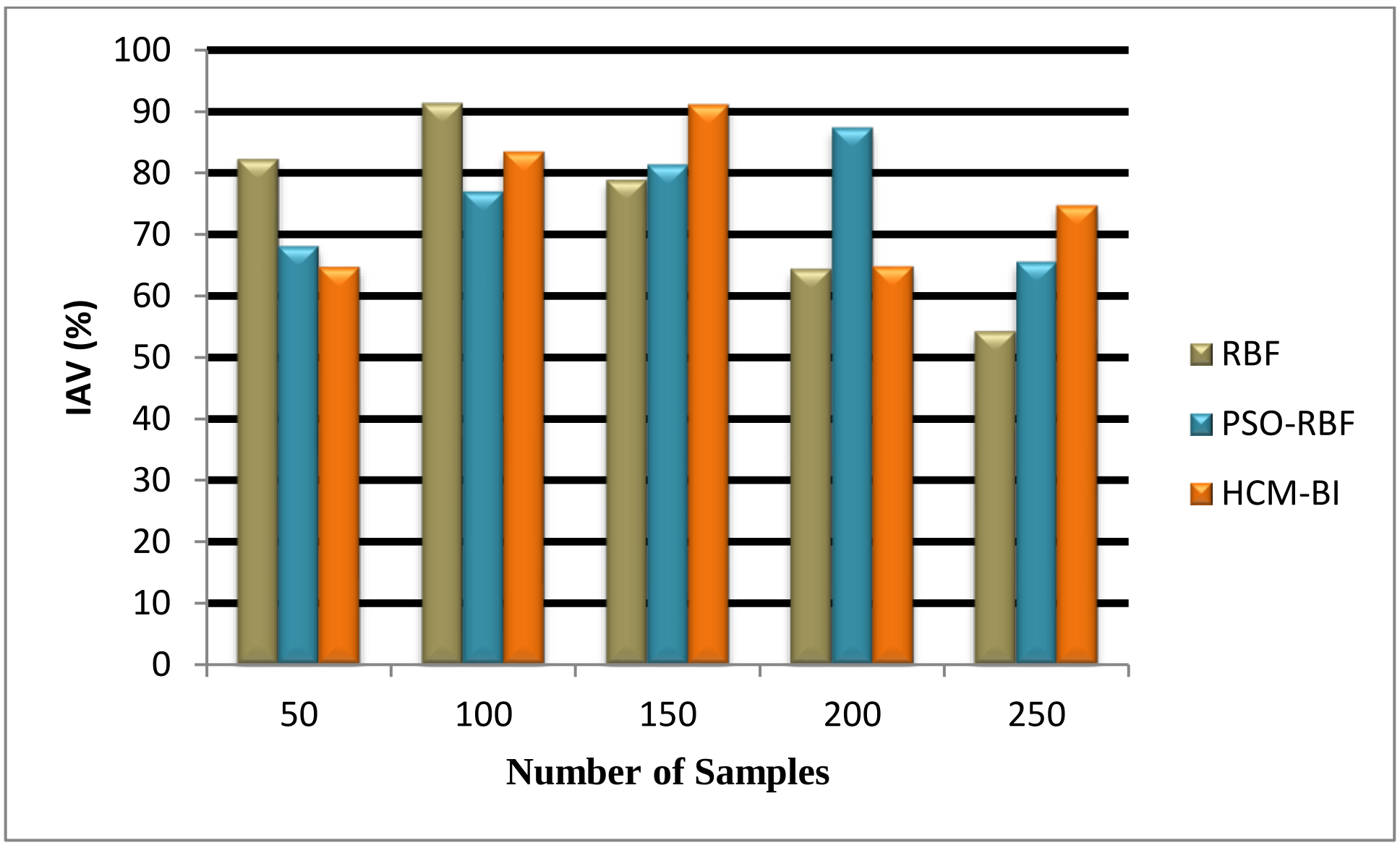

Figure11: Integral Absolute Value of a particular attribute

The log evidence is obtained by Equation (13). Depending upon the number of nodes, the evidence rate of hybridized classification model for Hypoxia-Ischemia and Hypoglycemia, Epilepsy brain injury (HCM-BI) is achieved.

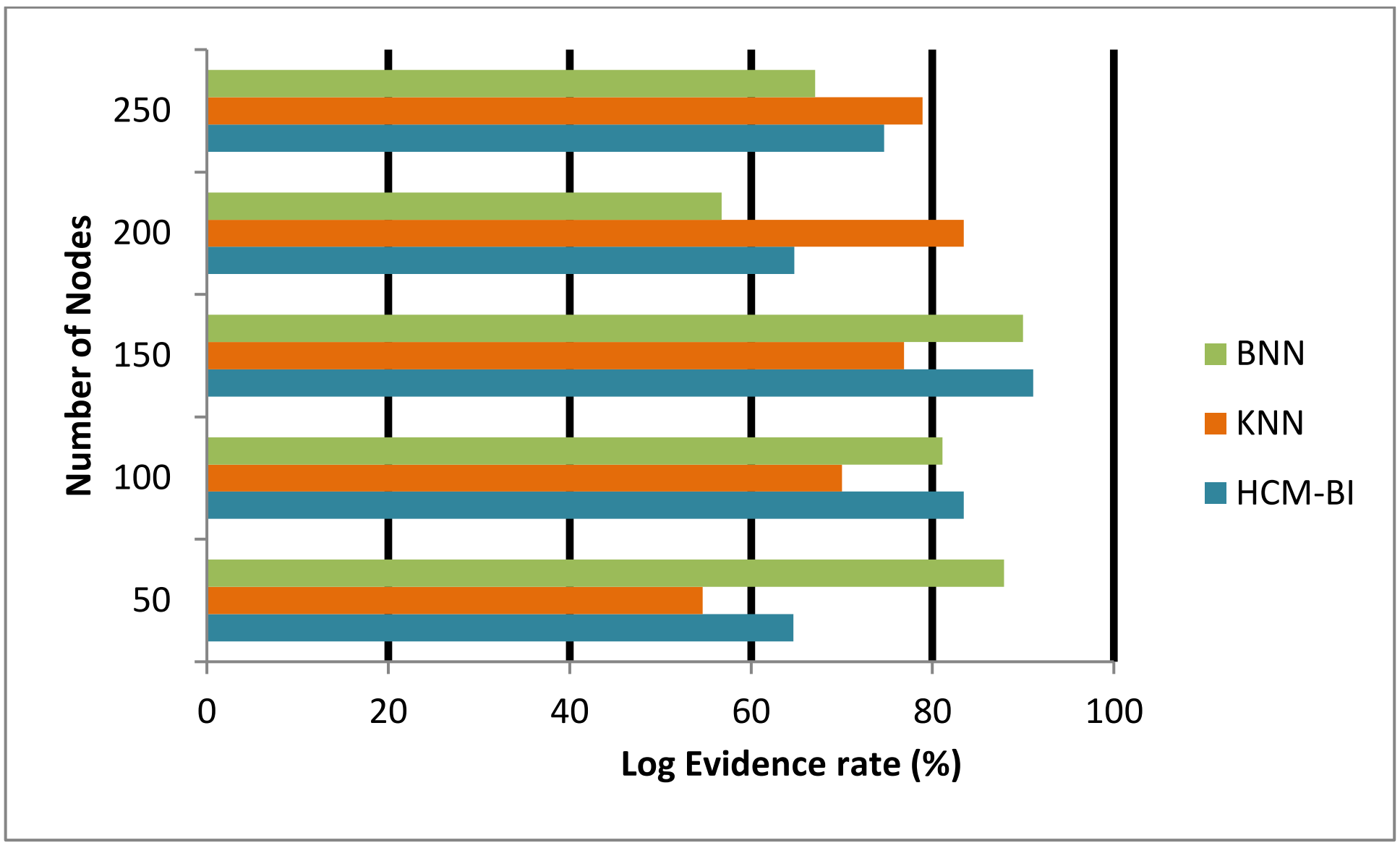

Figure12: Log Evidence by support vector machine

As inferred from the figure.12 The classification accuracy is compared by taking the dataset from the hospital, and this dataset has five divisions (A, B, C, D, E), in this A and B are the data taken from normal and healthy people C, D, E are collected from the brain-injured persons. Depending upon the dataset sample, classification accuracy varies, as shown in Table 1. The main aim would be to find a set of extracted features for each sample. In the precision of classification in the light of EEG feature vectors, four typical attributes are involved in this stage that can be denoted as $[g_1, g_2, g_3, g_4]$ helps to reduce the accuracy.

Table 1: A/E Classification Accuracy

| Number of Iterations | K-Nearest Neighbor | SVM | RBF | HCM-BI |
|---|---|---|---|---|
| 10 | 62.41 | 59.60 | 66.88 | 75.66 |
| 20 | 43.21 | 71.34 | 72.99 | 80.34 |
| 30 | 71.12 | 62.94 | 48.87 | 91.44 |
| 40 | 30.10 | 71.23 | 55.78 | 88.78 |
| 50 | 76.89 | 49.99 | 62.45 | 90.66 |

A/E classification accuracy is the sample of normal and fully affected brain-injured persons. The samples of the person are compared in the dataset, and the accuracy is defined. CD/E classification accuracy is shown in table 2. CD represents the slightly affected by brain injury, and E represents the completely affected person

Table 2: CD/E Classification Accuracy

| Number of Iterations | K-Nearest Neighbor | SVM | RBF | HCM-BI | |
|---|---|---|---|---|---|
| 10 | 67.88 | 57.89 | 81.23 | 87.99 | |
| 20 | 56.77 | 63.01 | 78.99 | 89.98 | |
| 30 | 45.88 | 52.11 | 76.09 | 91.44 | |
| 40 | 78.65 | 71.89 | 81.11 | 94.66 | |
| 50 | 56.98 | 70.22 | 85.66 | 95.5 | |

Based on the system structure, a support vector machine is employed with clinical information to determine each infant's HI outcomes using a hybridized classification model for hypoxia-ischemia and hypoglycemia, epilepsy brain injury (HCM-BI). According to the experimental findings, the proposed strategy improves accuracy by 95.05% and reduces error rates for illness comparisons to 0.41.

## 5. Conclusion

This research provides information regarding Hypoxia-Ischemia (HI), Hypoglycemia, and Epilepsy damage to the brain for an extended period. In the current situation, the potential to combine medical, Electroencephalography (EEG) measurements to forecast outcomes in 2 years is examined. HCM-BI defines the HI outcomes of each infant; the support vector machine is applied with clinical details. A selection of four attributes is derived from the Electroencephalography records, and SVM does not get clear results regarding the classification of diseases. The Bayesian neural network (BNN) optimizes the final feature extraction of the EEG signal to obtain the precise health status of patients with hypoglycemia and epilepsy. The experimental findings demonstrate that the suggested strategy improves accuracy by 95.05% and reduces the error rate to 0.41 when comparing diseases.

**Funding**: "This research received no external funding"

**Conflicts of Interest**: "The authors declare no conflict of interest."